%% file: header.tex
\documentclass[sigconf]{acmart}
\renewcommand\footnotetextcopyrightpermission[1]{} % removes footnote with conference information in first column
\setcopyright{none}
\def\BibTeX{{\rm B\kern-.05em{\sc i\kern-.025em b}\kern-.08emT\kern-.1667em\lower.7ex\hbox{E}\kern-.125emX}}

\usepackage{multirow}
\usepackage{array}
\usepackage{siunitx}
\usepackage{nicefrac}
\usepackage[utf8]{inputenc}
\usepackage{hyperref}
\hypersetup{
    bookmarks=true,         % show bookmarks bar?
    unicode=false,          % non-Latin characters in Acrobat’s bookmarks
    pdftoolbar=true,        % show Acrobat’s toolbar?
    pdfmenubar=true,        % show Acrobat’s menu?
    pdffitwindow=false,     % window fit to page when opened
    pdfstartview={FitH},    % fits the width of the page to the window
    pdftitle={My title},    % title
    pdfauthor={Author},     % author
    pdfsubject={Subject},   % subject of the document
    pdfcreator={Creator},   % creator of the document
    pdfproducer={Producer}, % producer of the document
    pdfkeywords={keyword1, key2, key3}, % list of keywords
    pdfnewwindow=true,      % links in new PDF window
    colorlinks=false,       % false: boxed links; true: colored links
    linkcolor=red,          % color of internal links (change box color with linkbordercolor)
    citecolor=green,        % color of links to bibliography
    filecolor=magenta,      % color of file links
    urlcolor=cyan           % color of external links
}

\usepackage{graphicx}
\usepackage{mwe}
\usepackage{subcaption}
\newcolumntype{L}[1]{>{\raggedright\let\newline\\\arraybackslash\hspace{0pt}}m{#1}}
\newcolumntype{C}[1]{>{\centering\let\newline\\\arraybackslash\hspace{0pt}}m{#1}}
\newcolumntype{R}[1]{>{\raggedleft\let\newline\\\arraybackslash\hspace{0pt}}m{#1}}

\begin{document}

\title[Multimodal Analytics for Real-world News using Measures of Cross-modal Entity Consistency]{Multimodal Analytics for Real-world News using \\ Measures of Cross-modal Entity Consistency}

\author{Eric M\"uller-Budack}
\email{eric.mueller@tib.eu}
\orcid{0000-0002-6802-1241}
\affiliation{%
 \institution{Leibniz Information Centre for Science and Technology~(TIB)}
 \city{Hannover}
 \state{Germany}
}

\author{Jonas Theiner}
\email{theiner@stud.uni-hannover.de}
\affiliation{%
 \institution{Leibniz Universit\"at Hannover}
 \city{Hannover}
 \state{Germany}
}

\author{Sebastian Diering}
\email{diering@stud.uni-hannover.de}
\affiliation{%
 \institution{Leibniz Universit\"at Hannover}
 \city{Hannover}
 \state{Germany}
}

\author{Maximilian Idahl}
\email{idahl@l3s.de}
\affiliation{%
 \institution{L3S Research Center, Leibniz Universit\"at Hannover}
 \city{Hannover}
 \state{Germany}
}

\author{Ralph Ewerth}
\email{ralph.ewerth@tib.eu}
\orcid{0000-0003-0918-6297}
\affiliation{%
 \institution{Leibniz Information Centre for Science and Technology~(TIB)}
 \institution{L3S Research Center, Leibniz Universit\"at Hannover} 
 % \city{Hannover}
 % \state{Germany}
}

% \renewcommand{\shortauthors}{Eric M\"uller-Budack, Jonas Theiner, Sebastian Diering, Maximilian Idahl, Ralph Ewerth}

\begin{abstract}
The World Wide Web has become a popular source for gathering information and news. Multimodal information, e.g., enriching text with photos, is typically used to convey the news more effectively or to attract attention. Photo content can range from decorative, depict additional important  information, or can even contain misleading information. Therefore, automatic approaches to quantify cross-modal consistency of entity representation can support human assessors to evaluate the overall multimodal message, for instance, with regard to bias or sentiment. In some cases such measures could give hints to detect fake news, which is an increasingly important topic in today's society. 
In this paper, we introduce a novel task of cross-modal consistency verification in \textit{real-world news} and present a multimodal approach to quantify the entity coherence between image and text. Named entity linking is applied to extract persons, locations, and events from news texts. Several measures are suggested to calculate cross-modal similarity for these entities using state of the art approaches. 
In contrast to previous work, our system automatically gathers example data from the Web and is applicable to real-world news. Results on two novel datasets that cover different languages, topics, and domains demonstrate the feasibility of our approach. Datasets and code are publicly available\footnote{\url{https://github.com/TIBHannover/cross-modal_entity_consistency}\label{foot:github}} to foster research towards this new direction.
\end{abstract}


%   \includegraphics[width=\textwidth]{graphics/examples_with_text_2.pdf}

\maketitle
%
%
%
\input{content.tex}
\bibliographystyle{ACM-Reference-Format}
\bibliography{references}

\end{document}

%% file: content.tex
% #############################################################################
\section{Introduction}
\label{chp:intro}
% #############################################################################
%
% ---- Motivation ----
In today's information age, the Web plays an important role in disseminating information and news. Social media in particular can quickly inform users about worldwide events and have become a popular news source.
\input{fig_intro}
These news articles often leverage different modalities, e.g. texts and images, to communicate information more effectively~(Figure~\ref{fig:examples}). Photo content can range from decorative~(with little or no information about the news event) over depicting rich information enhancements~(showing important or additional details) to even misleading visual information. Therefore, automatic approaches to quantify the cross-modal consistency of entity representations can support human assessors to evaluate the overall multimodal message, the cross-modal presence of entities, or search in large multimodal news corpora. In some cases measures of cross-modal entity consistency might also help to identify fake news, i.e., articles that deliberately spread rumours or misleading information. But the problem is manifold and solutions of different topics such as image repurposing detection~\cite{jaiswal2017multimedia,sabir2018deep,jaiswal2019aird}, text-based rumor detection~\cite{gupta2012evaluating,liu2015real,ma2016detecting}, and image forensics\cite{bondi2017first,huh2018fighting,luo2010jpeg,popescu2005exposing,salloum2018image,wu2017deep,zhou2018learning} need to be combined to support users or expert-oriented fact checking efforts such as \mbox{\emph{PolitiFact}} and \emph{Snopes} to reveal disinformation. 
%
% Related Work
%
While part of related work~\cite{henning2018estimating,otto2019understanding} aims to find measures to model semantic cross-modal relations in order to bridge the semantic gap, approaches on image repurposing detection~\cite{jaiswal2017multimedia,sabir2018deep,jaiswal2019aird} are suggested to check the consistency of factual information, or more specifically named entities~(as illustrated in~Figure~\ref{fig:examples}). 
This work is similar to approaches for image repurposing and focuses on measures to evaluate the consistency of named entities between image and text. Related approaches~\cite{jaiswal2017multimedia,jaiswal2019aird,sabir2018deep} rely on multimodal deep learning techniques that require appropriate training or reference datasets. These datasets contain non-manipulated pairs of image and text that need to be verified for valid cross-modal relations, which makes it hard to collect them automatically. In addition, training or reference data are crucial for system performance since they provide the source of world knowledge. Thus, these methods are restricted to entities, e.g., persons or locations, that appear in these datasets, which is a severe limitation for real-world scenarios. Experimental evaluation has been performed on images from Flickr or public benchmarks with rather short image captions~\cite{jaiswal2017multimedia,sabir2018deep} or existing metadata~\cite{sabir2018deep}, which do not reflect real-world characteristics as illustrated in Figure~\ref{fig:examples}.

% ---- Contributions ---- 
In this paper, we present a fully-automatic system that aims to support human assessors with valuable measures of cross-modal entity consistency. In contrast to previous work, the system is completely unsupervised and does not rely on any predefined reference or training data. To the best of our knowledge, we present a first baseline that is \emph{applicable to real-world news articles} by tackling several news-specific challenges such as the excessive length of news documents, entity diversity, and noisy reference images. More specifically, we automatically crawl reference images for entities that are extracted from the text using a named entity linking approach. These images serve as input for the visual verification of the entities to the accompanying news image and appropriate computer vision approaches serve as generalized feature extractors. Finally, novel measures for different entity types~(person, location, event) as well as for a more general news context are introduced to quantify the cross-modal similarity of image and text. The applications for our proposed approach are manifold, ranging from a retrieval system for news with low or high cross-modal coherence to an exploration tool that reveals the relations between image and text.
%
% ---- Evaluation ----
The feasibility of our approach is demonstrated on a novel large-scale dataset for cross-modal consistency verification that is derived from \emph{BreakingNews}~\cite{ramisa2018breakingnews}. The dataset contains real-world news in English and covers different topics and domains. In addition, we have collected articles from \emph{German} news sites to verify the performance in another language. In contrast to previous work, the entities are tampered with more sophisticated strategies to obtain challenging datasets. Source code, web application, and datasets are publicly available\textsuperscript{\ref{foot:github}} to foster research in this new direction. 

% ---- Rest of the Paper ----
The remainder of the paper is organized as follows. In Section~\ref{chp:rw}, we review related work and focus on multimodal image repurposing detection. Our framework to automatically verify cross-modal relations is described in Section~\ref{chp:methodology}. Experimental results for two different datasets are presented and discussed in Section~\ref{chp:exp}. Section~\ref{chp:summary} summarizes the paper and outlines areas of future work.
%
% #############################################################################
\section{Related Work}
\label{chp:rw}
% #############################################################################
%
% ---- Solely Text or Image ----
The increasing importance of fake news and rumor detection has gained some interest in computer science research recently. Most of the proposed approaches are based on the textual content~\cite{gupta2012evaluating,liu2015real,ma2016detecting} in combination with supplementary information such as the responses of other people~\cite{ruchansky2017csi}, or dissemination characteristics of the article~\cite{jin2013epidemiological,wu2015false}. Visual forensic methods~\cite{bondi2017first,huh2018fighting,luo2010jpeg,popescu2005exposing,salloum2018image,wu2017deep,zhou2018learning} aim to detect forged images and can also be used as a potential indicator for fake news. Some approaches~\cite{gupta2013faking,jin2017novel} directly exploit visual statistics of untampered images used in real and fake news articles to add valuable features to reveal fake news. 
%
% ---- Multimodal ----
%
However, another important clue is the multimodal consistency of image and text. While few approaches~\cite{henning2018estimating,otto2019understanding} explore more general semantic correlations between image and text, related work on image repurposing~\cite{jaiswal2017multimedia,sabir2018deep,jaiswal2019aird} is more similar to our approach and focuses on the verification of specific meta information. \citet{jaiswal2017multimedia} learn a multimodal representation of reference packages that contain an untampered image and a corresponding caption to assess the semantic integrity of a given document. Experiments were conducted by replacing one modality which results in semantically inconsistent image-captions pairs and thus making them relatively easy to detect. This motivated \citet{sabir2018deep} to introduce a dataset where specific entities~(persons, locations, and organizations) are carefully replaced to generate semantically consistent altered packages. They have also refined the multimodal model using a multitask learning approach that further incorporates geographical information. \citet{jaiswal2019aird} presented an adversarial neural network that simultaneously trains a bad actor who intentionally counterfeits metadata and a watchdog who verifies the multimodal semantic consistency. 
However, these approaches neglect the various challenges of real-world news and applications in terms of the huge amount and variety of entities, noisy reference data and noisy outputs of named entity linking tools. They instead rely on pre-defined reference datasets consisting of image-text pairs~\cite{jaiswal2017multimedia,sabir2018deep}, or existing metadata~\cite{jaiswal2019aird} that are (1)~closely related~(Figure~\ref{fig:examples} top), (2)~hard to collect automatically, and (3)~rather limited and static with respect to the covered entities.
%
% #############################################################################
\section{Cross-modal Consistency Verification}
\label{chp:methodology}
\begin{figure*}[t]
	\centering
    \includegraphics[width=0.93\linewidth]{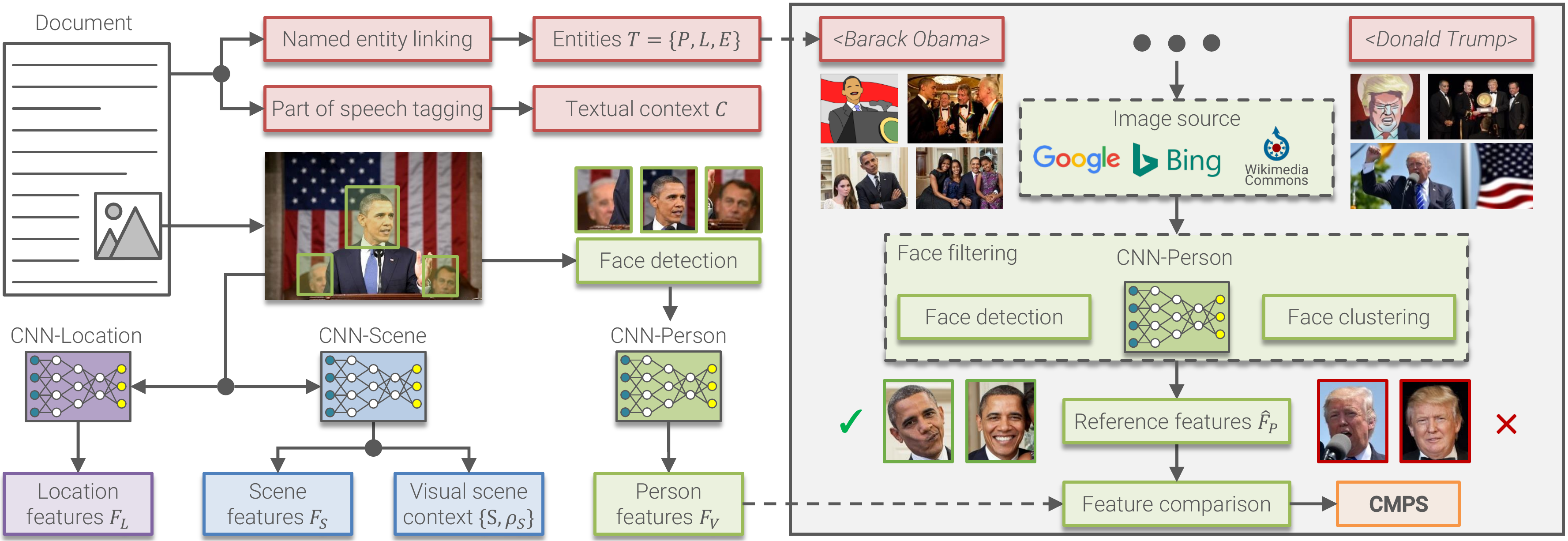}
	\caption{Workflow of the proposed system. Left: Extraction of textual entities~$T$ and context~$C$~(red) according to Section~\ref{sec:txt_features}, as well as visual features~$F$~(Section~\ref{sec:vis_features}) for persons~(green), locations~(purple), and scenes~(blue). In addition, the visual scene context containing the probabilities~$\rho_S$ and word embeddings~$S$ of all scene classes is calculated. Right: Workflow to measure the \emph{Cross-modal Person Similarity} between image and text. A similar pipeline is used for locations and events~(Section~\ref{sec:align_features}).}
	\label{fig:feature_extraction}
\end{figure*}

% #############################################################################
%
In this section, we present the proposed unsupervised system that automatically verifies the semantic coherence between pairs of image and text. Verification is realized and operationalized through system outputs of measures of cross-modal similarities for different entity types~(persons, locations, and events) as well as a more general context. % We show how to overcome several challenges in real-world news. 
Based on named entity linking~(Section~\ref{sec:txt_features}) visual evidence for the cross-modal occurrence of entities is gathered from the Web. Appropriate computer vision approaches are applied to obtain rich feature representations~(Section~\ref{sec:vis_features}), which are used in conjunction with measures of cross-modal similarity~(Section~\ref{sec:align_features}) to quantify the cross-modal consistency. The workflow is illustrated in Figure~\ref{fig:feature_extraction}. 
%
%
% #####################################
\subsection{Extraction of Textual Entities and Context}
\label{sec:txt_features}
% #####################################
%
% First, named entities as well as contextual candidates are extracted from the text as explained in the following Section.
%
\subsubsection{Persons, Locations, and Events:}
\label{sec:txt_entities}
% TODO wikifier huge amount of supported languages!
We have tried several frameworks for named entity linking such as \emph{AIDA}~\cite{hoffart2011robust}, \emph{NERD}~\cite{rizzo2012nerd}, or \citet{kolitsas2018end}'s approach. However, we found that combining the output of \emph{spaCy}~\cite{honnibal2017spacy} and \emph{Wikifier}~\cite{brank2018semantic} provides the best results for the detection and disambiguation of named entities for different languages. It also enables our system to support a large number of 100~languages. We link the entity candidate with the highest \emph{PageRank} according to \emph{Wikifier} for every named entity recognized by \emph{spaCy} to the \emph{Wikidata} knowledge base. If \emph{Wikifier} does not provide any linked entity for a given string, the \emph{Wikidata}~API is utilized for disambiguation. 
To handle mistakes regarding the entity type classification by \emph{spaCy} and to discard irrelevant entities such as given names, we re-evaluated the entity type by defining the following requirements for persons~$P$, locations~$L$, and events~$E$. For persons only entities that are an instance of~\emph{human} according to \emph{Wikidata} are considered, while for locations a valid \emph{coordinate location} is set as a requirement. This allows us to extract a variety of locations ranging from continents, countries, and cities to specific landmarks, streets or buildings. For events we instead require an entity to be in a verified list of events~\cite{gottschalkeventlist} according to \emph{EventKG}~\cite{gottschalk2018eventkg}.

\subsubsection{Textual Scene Context:}
\label{sec:txt_embedding}
To retrieve candidates representing the (scene)~context~$C$ of the text, the part-of-speech tagging from \emph{spaCy}~\cite{honnibal2017spacy} is applied to extract all nouns. They can contain general concepts, such as politics or sports, as well as specific scenes or actions. Subsequently, we calculate the word embedding for each candidate using \emph{fastText}~\cite{grave2018learning} as a prerequisite for the cross-modal comparison explained in Section~\ref{sec:align_features}.
%
% #####################################
\subsection{Extraction of Visual Features}
\label{sec:vis_features}
% #####################################
%
Our approach is applicable to articles that contain multiple images, but for simplification we assume that only a single image is present in a document. For person verification, we first jointly detect and normalize faces using the \emph{Multi-task Cascaded Convolutional Networks}~\cite{zhang2016joint}. State of the art models are applied to obtain rich visual image representations. 

% PERSONS & LOCATIONS
An implementation~\cite{sandbergfacenet} of \emph{FaceNet}~\cite{schroff2015facenet} is used to calculate the feature vectors of each face found in the image. To get a geospatial representation of the whole image, we employ the \emph{base}\,($M,f\text{*}$) model~\cite{muller2018geolocation} for geolocalization since it provides good results across different scene types~(indoor, natural, urban). In contrast to the original method, we treat geolocalization as a verification approach and utilize the feature vector from the penultimate pooling layer of the model.
%
% EVENTS & CONTEXT
We are not aware of a computer vision model for event classification that is capable to distinguish between the majority of real-world event types such as elections or natural disasters. Thus, a more general image descriptor for scene classification~\cite{zhou2017places} is applied. The visual scene features from the last pooling layer of a \emph{ResNet} model~\cite{resnetplaces,he2016deep} serve as an input for event verification.
Furthermore, the classification output is calculated to retrieve the probabilities of the $365$~scene concepts in the \emph{Places2} dataset~\cite{zhou2017places}. As for the textual scene context described in Section~\ref{sec:txt_features}, \emph{fastText}~\cite{grave2018learning} is employed to extract the corresponding word embeddings of the scene class. Both the scene probabilities~$\rho_S$ and word embeddings~$S$ are used as visual scene context. We also manually translated the labels to \emph{German} to enable experiments with news documents in another language.
%
% #####################################
\subsection{Verification of Shared Cross-modal Entities}
\label{sec:align_features}
% #####################################
%
In this section, we explain how to measure the cross-modal similarity of persons, locations, and events. Furthermore, the scene context is compared between image and text for a more general contextual cross-modal similarity. 
It should be emphasized that we treat each verification task independently. The cross-modal similarities for different entity types are \emph{not combined} which allows a more detailed and realistic analysis. Referring to Figure~\ref{fig:examples}~(bottom), please imagine a news article where the image depicts one ore multiple person(s) talking at a conference. While there can be multiple events and locations mentioned in the corresponding text, the news image does not provide any visual cues for their verification. This is very typical in news articles since the text usually contains more entities and information. In case of fake news commonly also only one entity type is tampered to maintain credibility.
%
% #################
\subsubsection{Verification of Persons:}
\label{sec:align_persons}
% #################
%
As illustrated in Figure~\ref{fig:feature_extraction}, we first gather a maximum of $k$~example images using image search engines such as \emph{Google} or \emph{Bing} for each person~$p \in P$ that was extracted from the named entity linking approach presented in Section~\ref{sec:txt_entities}. Since these images can contain noise or depict multiple persons, a filtering step is necessary. As in Section~\ref{sec:vis_features} feature vectors are extracted for each face detected in the images. These features are compared with each other using the cosine similarity to perform a hierarchical clustering with a minimal similarity threshold~$\tau_P$ as a termination criterion. Consequently, the mean feature vector of the majority cluster is calculated and serves as the reference vector~$\hat{F}_p$ for person~$p$, since it most likely represents the queried person.

Finally, the feature vectors~$F_V$ of all faces~$v \in V$ in the document image are compared to the reference vectors~$\hat{F}_P$ of each person~$p \in P$ mentioned in the text. Several options are available to calculate an overall \emph{Cross-modal Person Similarity}~(CMPS) such as the mean, \mbox{$n\%$-quantile}, or the max of all comparisons. However, as mentioned above, usually the text contains more entities than the image and already a single correlation can theoretically ensure the credibility. Thus, we define the \emph{Cross-modal Person Similarity}~(CMPS) as the maximum similarity among all comparisons according to Equation~\ref{eq:CMPS}, since the mean or quantile would require the presence of several or all entities mentioned in the text. 
\begin{equation}
\label{eq:CMPS}
    \textnormal{CMPS} = \max_{v \in V, p \in P}{\left(\frac{F_v \cdot \hat{F}_p}{||F_v|| \cdot ||\hat{F}_p||}\right)}
\end{equation}
%
% #################
\subsubsection{Verification of Locations and Events:}
\label{sec:align_locations}
% #################
% 
In general, we follow the pipeline of person entity verification. The feature vectors of the reference images for each location and event mentioned in the text are calculated using the CNN of the respective entity type. But while some entities are very specific~(e.g., landmarks, sport finals), others are more general~(e.g., countries, international crises) and can therefore contain diverse example data. This makes a visual filtering based on clustering very complicated as these entities can already contain many visually different subclusters. Thus, we compare the feature vector of the news photo to the feature vector of each reference image using the cosine similarity. To obtain the cross-modal similarity for each entity, the maximum and several n\%-quantiles are evaluated in the experiments~(Section \ref{sec:params}). We believe that using a n\%-quantile is more robust against noise in the retrieved reference material. As explained for person verification, we decided to use the maximum cross-modal similarity among all entities of a given type for both the \emph{Cross-modal Location Similarity}~(CMLS) and \emph{Cross-modal Event Similarity}~(CMES) of the document.
%
% #################
\subsubsection{Scene Context Verification:}
\label{sec:align_scenes}
% #################
%
The \emph{Cross-modal Context Similarity}~(CMCS) is calculated based on a comparison of word embeddings and does not require any reference images. The cosine similarity of the word embedding~$C$ of the textual context candidates~(Section~\ref{sec:txt_embedding}) to the embeddings~$S$ of the $365$~scene classes in the Places2 dataset~\cite{zhou2017places}~(Section~\ref{sec:vis_features}) is calculated. Since only certain scenes are represented in the news image, these similarities are weighted with the respective visual scene probability~$\rho(s)$. The CMCS is defined as the maximum similarity of all comparisons:
\begin{equation}
\label{eq:CMCS}
    \textnormal{CMCS} = \max_{c \in C}\left(\sum_{s \in S}{\rho(s) \cdot \frac{s \cdot c}{||s|| \cdot ||c||}}\right)
\end{equation}
%
% #############################################################################
\section{Experimental Setup and Results}
\label{chp:exp}
% #############################################################################
%
In this section, we introduce two novel datasets for cross-modal consistency verification~(Section~\ref{sec:datasets}). In addition, the metrics for evaluation~(Section~\ref{sec:metrics}) and parameter selection~(Section~\ref{sec:params}) are explained in more detail. Finally, the performance of the proposed system on real-world news articles~(Section~\ref{sec:results}) is discussed.
\subsection{Datasets \& Setup}
\label{sec:datasets}
\input{tab_statistics_datasets}
%
Two real-world news datasets that cover different languages, domains, and topics are utilized for the experiments~(Table~\ref{tab:statistics_datasets}). They were both manipulated to perform experiments for cross-modal consistency verification. Experiments and comparisons to related work~\cite{jaiswal2017multimedia,sabir2018deep} on datasets such as \emph{MEIR}~\cite{sabir2018deep} are not reasonable since 1) they do not contain public persons or events, and  2) rely on \textit{pre-defined} reference or training data for \textit{given} entities. These restrictions severely limit the application in practice. We propose an automated solution for real-world scenarios that works for public personalities and entities represented in a knowledge base. Source code and datasets to reproduce our results are publicly available\textsuperscript{\ref{foot:github}}.
\subsubsection{Tampering Techniques:}
\label{sec:tampering}
We have created multiple sets of tampered entities for each document in our datasets. Similar to \citet{sabir2018deep}, we replaced entities extracted from the text at random with another entity of the same type to maintain the semantic coherence. We also apply more sophisticated tampering techniques as follows.
Three additional tampered person sets are created by replacing each untampered person with another person of the same gender~(PsG), same country of citizenship~(PsC), or matching both aforementioned criteria~(PsCG). 
Locations are replaced by other locations that share at least one parent class~(e.g., country or city) according to \emph{Wikidata} and are located within a Great Circle Distance~(GCD(${dmin}, {dmax}$)) of~$dmin$ and~$dmax$ kilometers. Three intervals are used to experiment with different spatial resolutions. 
Similarly, events that share the same parent class~(e.g., sport competition or natural disaster) with the untampered event are used for a second set~(EsP) of tampered events. In case no valid candidate for a tampering strategy was available, we have used a random candidate that matched the most other tampering criteria. 

The contextual verification is based on the nouns in the text. Thus, textual tampering techniques are not applicable. We instead replaced the image with a random image from all other documents for a first tampered set. We randomly selected similar images~(from top-$k\%$ with $k \in \{5, 10, 25\}$) to maintain semantic coherence to create three more sets. The similarity was computed using feature vectors extracted from a \emph{ResNet} model~\cite{he2016deep} trained on \emph{ImageNet}~\cite{deng2009imagenet}.
\subsubsection{TamperedNews Dataset:}
\label{sce:breaking}
To the best of our knowledge, \emph{BreakingNews}~\cite{ramisa2018breakingnews} is the largest available corpus with news articles that contain both image and text. It originally covered approximately $100,000$~news articles published in 2014 written in English across different domains and a huge variety of topics~(e.g., sports, politics, healthcare). We created a subset~(\emph{TamperedNews}) for cross-modal consistency verification of $72,561$~articles for which the news text and image were still available. The entities in these articles were additionally tampered according to~Section~\ref{sec:tampering}. To discard most irrelevant entities, only persons and locations that are mentioned at least in ten as well as events that occur in at least three documents are considered. Detailed dataset statistics are reported in Table~\ref{tab:statistics_datasets}. 
\subsubsection{News400 Dataset:}
\label{sec:news400}
To show the capability of our approach for another language, we have used the \emph{Twitter API} to obtain the web links~(URLs) of news articles from three popular \emph{German} news websites~(\url{faz.net}, \url{haz.de}, \url{sueddeutsche.de}). The texts and main images of the articles were crawled from the URLs. We have gathered $400$~news articles containing four different topics~(\emph{politics, economy, sports, and travel}) in the period from August~2018 to January~2019. The smaller size of the dataset allowed us to conduct a manual annotation with three experts to ensure valid relationships between image and text. 
For each document, the annotators verified the presence of at least one person, location, or event in the image as well as in the text, and whether the context was consistent in both modalities. Experiments were conducted exclusively on data with valid relations. Again the tampering techniques presented in Section~\ref{sec:tampering} are applied to create the test sets. Due to its smaller size, every entity is considered regardless of how often it appears in the entire dataset. The resulting statistics are shown in Table~\ref{tab:statistics_datasets}.
\subsection{Evaluation Tasks \& Metrics}
\label{sec:metrics}
The evaluation tasks are motivated by potential real-world applications of our system. We propose to evaluate the system for two tasks: (1)~document verification and (2)~collection retrieval. The system can also be used as an analytics tool to quickly explore cross-modal relations within a document as illustrated in Figure~\ref{fig:examples}. 

\textbf{Document Verification:} Please imagine a set of two or more news articles with similar content and imagery but differences in the mentioned entities that might have been tampered by an author with harmful intents. The idea behind this task is to decide which joint pair of image and entities extracted from the news text provides a higher cross-modal consistency. Thus, a document verification can help users to detect the most or least suitable document. We address this task using the following strategy.
For each individual document in the dataset we compare the cross-modal similarities between the news image and the respective set of untampered entities as well as \emph{one} set of tampered entities~(e.g, PsG) according to the strategies proposed in Section~\ref{sec:tampering}. This allows us to evaluate the impact of different tampering strategies. We report the \emph{Verification Accuracy}~(VA) that quantifies how often the untampered entity set has achieved the higher cross-modal similarity to the document's image. Some qualitative examples are shown in Section~\ref{sec:results_tamperednews}~(Figure~\ref{fig:case_study}). Please note that for the context evaluation the image is tampered instead and that the nouns in the text are considered as "entities". As mentioned at the beginning of Section~\ref{sec:align_features}, we only consider entities of one type~(e.g, persons) for the respective experiments for a more detailed and realistic analysis.

\textbf{Collection Retrieval:} The system can also be leveraged in news collections to retrieve news articles with high or low cross-modal relations in order to support human assessors to gather the most credible news or possibly fake news~(in extreme cases). In contrast to document verification, we therefore consider all $|D|$~untampered documents as well as $|D|$~tampered documents applying \emph{one} tampering strategy. The cross-modal similarities are calculated and used to rank all $2 \cdot |D|$~documents. As suggested by previous work~\cite{jaiswal2017multimedia,sabir2018deep}, AUC~(Area Under Receiver Operating Curve) is used for evaluation. In addition, we propose to calculate the average precision for retrieving untampered~(AP-clean) or tampered~(AP-tampered) documents at specific recall levels~$R$ according to Equation~(\ref{eq:AP}). In this respect, $TP^i$ is the number of relevant documents at position~$i$. For example, AP-tampered@25\% describes the average precision when $|D_R| = 0.25 \cdot |D|$ of all tampered documents are retrieved.
\begin{equation}
\label{eq:AP}
    AP@R = \frac{1}{|D_R|} \sum_{i=1}^{k}{\frac{TP^i}{i}} \; ,
\end{equation}

\textbf{Test Document Selection for TamperedNews:} Although, the large size of the \emph{TamperedNews} dataset allows for a large-scale analysis of the results, unfortunately a manual verification of cross-modal relations as for \emph{News400} is infeasible. Thus, calculating the proposed metrics for the whole dataset can lead to misleading results since during the annotation of \emph{News400} it turned out that only a fraction of the documents have shown cross-modal entity correlations~(Table~\ref{tab:statistics_datasets}). As discussed at the beginning of Section~\ref{sec:align_features}, it is possible that not a single entity mentioned in a news text is depicted in the corresponding image. To address this issue, we suggest to measure the metrics for specific subsets. More specifically, we consider the top-$25\%$ and top-$50\%$ documents~(denoted as \emph{TamperedNews (Top-k\%)}) with respect to their cross-modal similarity of untampered entities since they more likely contain relations between image and text. This selection is also supported by the CMPS~values for person verification~(Figure~\ref{fig:cms}), which decrease more significantly after $25\% - 50\%$ of all documents and corresponds to the percentage of manually verified documents in the \emph{News400} dataset. 
\input{fig_cms_small}
\subsection{Parameter Selection}
\label{sec:params}
For parameter selection we have used 10\% randomly selected documents within each \emph{TamperedNews~(Top-k\%)} subset.
In total, we gathered a maximum of 20~images from the image search engines of \emph{Google} and \emph{Bing} as well as all~$k_{W}$ available images on \emph{Wikidata}~(mostly one \emph{Wikimedia} image) for each entity recognized in the text. We have used multiple sources to prevent possible selection biases of a specific image source and investigated the performance for different images sources and number of images. A detailed analysis goes beyond the scope of this work and is reported as supplemental material\textsuperscript{\ref{foot:github}}. 
% for the hardest tampering set and the AUC metric is presented in Table~\ref{tab:source_results}. All results can be found in the supplemental material allowing for similar conclusions. 
The results have demonstrated that the performance using a single or all image sources are very similar. In fact, results were sometimes a little worse when more images were used, which could indicate that the first images crawled from the search engines contain less noise. Hence, for the rest of our experiments we use all available image sources with a maximum of ten images per source, since this provides a good trade-off between performance and speed and prevents possible selection biases.

The threshold~$\tau_P$ has a significant impact on the agglomerative clustering approach to filter retrieved face candidates for a person~(as explained in Section~\ref{sec:align_persons}). For this reason, we have tested the \emph{FaceNet} model~\cite{schroff2015facenet} on the \emph{Labeled Faces in the Wild}~\cite{huang2008labeled} benchmark and evaluated an optimal cosine similarity~(normalized to the interval~[0, 1]) threshold of~$\tau_P=0.65$.

The results for different operators to combine the cross-modal similarities of all reference images to the news image for each entity~(as explained in Section~\ref{sec:align_features}) are presented in Table~\ref{tab:mode_results}.
\input{tab_mode_results.tex}
Surprisingly, results of 90\% and 95\% quantiles are on par with the proposed person clustering. Also contrary to our assumption that a quantile is more robust against noise for locations and events, it turned out that the maximum operator provides the best results for these entities. This indicates that noise in the reference data has no significant impact on the performance. Except for person entities, where reference faces can be very similar, we assume that noise less likely matches the entity depicted in the news image. In the remainder, results for persons are reported using the clustering strategy because we still believe that this is more robust in many scenarios. For locations and events the maximum operator is applied.

\subsection{Experimental Results}
\label{sec:results}
In this section, we present the baseline results of the proposed system for cross-modal consistency verification. Unfortunately, a comparison to previous work such as \citet{jaiswal2017multimedia} or \citet{sabir2018deep} is not reasonable, since these approaches can not handle the significantly longer news texts and need to be trained with labeled reference data that are much closer related to the source images. As discussed above, these approaches are not able do deal with real-world news in contrast to our approach.

\subsubsection{TamperedNews:} 
\label{sec:results_tamperednews}
\input{fig_results}
\input{tab_results_breakingnews}
Qualitative and quantitative results are presented in Figure~\ref{fig:cms}, Figure~\ref{fig:case_study}, and Table~\ref{tab:results_breakingnews}. Results for more document subsets~(e.g., \emph{TamperedNews~(Top-25\%)}) allow similar conclusions and is reported as supplemental material\textsuperscript{\ref{foot:github}}. 
As expected, the performance is best for person verification since the entities and the retrieved example material are very unambiguous and neural networks for face recognition, such as \emph{FaceNet}~\cite{schroff2015facenet}, can achieve impressive results. Despite the more challenging tampering techniques, our approach is still able to produce similar results. We have only experienced problems if persons were depicted in challenging conditions~(e.g., extreme poses as shown in Figure~\ref{fig:case_study}a for \emph{John Kerry}) or were rather unknown, which results in false entity linking results and confusion with other persons~(e.g., with a similar name).  

% --- Geolocation Results ----
To evaluate performance for location verification, we distinguished between images of indoor and outdoor scenes using the scene classification approach~(explained in Section~\ref{sec:vis_features}) and the scene hierarchy of the \emph{Places2}~dataset~\cite{zhou2017places}. Due to the data diversity and ambiguity as well as the unequal distribution of photos on earth, geolocation estimation is a very complex problem that has attracted attention only in recent years~\cite{muller2018geolocation,vo2017revisiting,weyand2016planet}. Therefore, the results were expected to be worse with respect to the person verification. Despite the complexity good results were achieved for outdoor images, whereas the detection of modified indoor scenes is more challenging given the low amount of geographical cues and their ambiguity. However, even when on the harder tampering set with lower Great Circle Distances and/or similar appearance~(Figure~\ref{fig:case_study}b and d), the system is able to operate on a good level and shows promising results in particular for outdoor images.
In contrast to person entities, location entities are an instance of various parent classes such as \emph{countries} or \emph{cities}. For a more in depth-analysis we have calculated the results for all types of locations separately using the documents~$D_s$ where an instance of a given type has achieved the highest \emph{CMLS} within the untampered set of entities. 
\input{tab_meta_results}
The results for some location types are presented in Table~\ref{tab:meta_results}~(left) and show that the performance is best for more fine-grained entities such as \emph{tourist attractions} and \emph{cities}. The performance for coarse location types such as \emph{oceans}, \emph{mountain ranges}, and country \emph{states} are typically worse since they do not provide sufficient geographical cues or are too broad to retrieve suitable reference images. Despite the results for continents or countries are also comparatively high, we believe the reason is that the candidates for tampering are easier to distinguish since location of those types have higher geographical and cultural differences. For fine-grained entities the tampering is much more challenging as illustrated in Figure~\ref{fig:case_study}b and d.

% --- Event Results ----
Although a more general descriptor for scene classification is used for event verification, good results were achieved when tampering with random events. However, the results for EsP are noticeably worse. As for locations we have provided results of common event types in Table~\ref{tab:meta_results}~(right). While the results for \emph{festivals}, \emph{holiday}, and \emph{disasters} are promising, event types such as \emph{sports competitions}, \emph{awards}, and \emph{conventions} are much worse. We believe that this is caused by the high visual similarity of events within these types. For example, many news articles on \emph{sport competitions} in the \emph{TamperedNews} dataset are about football~(Figure~\ref{fig:case_study}e). The reference images of entities of this type often show typical scenes, e.g., players on the the football pitch. These scenes are very hard to distinguish for the scene classification network. To address this issue, an expert network specifically trained for event classification should be applied~(as suggested for persons and locations).

% --- Scene Context Results ----
The results for scene context verification indicate that our system is able to reliably detect documents with randomly changed images. But, as also stated by \citet{sabir2018deep}, this task is easy to solve since the semantic coherence is not maintained. When similar images are used for tampering this task becomes much harder. Since networks for object classification~(used for tampering) and scene classification~(used for verification) could produce comparable results, performance is steadily decreasing if the tampered image has a higher visual similarity and could even show the same scene, e.g., sport. However, our system is still able to hint towards cross-modal consistencies. 

\subsubsection{News400:}
\label{sec:results_news400}
\input{tab_results_news400}
Since the number of documents is rather limited and cross-modal mutual presence of entities was manually verified, results for \emph{News400} are reported for all documents with verified relations. Based on the results displayed in Table~\ref{tab:results_news400}, similar conclusions on the overall performance of the system can be drawn. However, results while retrieving tampered documents are noticeable worse. This is mainly caused by the fact that some untampered entities in the documents that are depicted in both image and text, can be either unspecific~(e.g. mentioning of a country) or the retrieved images for visual verification do not fit the document's image content. Since subsets of the top-k\% documents for \emph{TamperedNews} were used to counteract the influence of untampered documents that do not show any cross-modal relations~(as discussed in Section~\ref{sec:metrics}) this problem was bypassed. We have verified the same behavior for \emph{News400} when experimenting on these subsets. For more details, we refer to the supplemental material\textsuperscript{\ref{foot:github}}. In addition, performance for context verification is worse compared to \emph{TamperedNews}. We assume that this is due to the less powerful word embedding for the \emph{German} language.

Overall, the system achieves promising performance for cross-modal consistency verification of persons, spatial information~(location), and spatio-temporal information in form of events. Since it dynamically gathers example data from the Web it is robust to changes in topics and entities, even when applied on news articles from another country and publication date. 
%
% #############################################################################
\section{Conclusions}
\label{chp:summary}
% #############################################################################
%
In this paper, we have presented a novel analytics system and benchmark datasets to measure the cross-modal consistency in real-world news articles. Named entity recognition is applied to find persons, locations, and events in the textual content to define queries using a knowledge base. These queries are utilized to automatically gather reference data, i.e., example images, for the visual verification of entities in the article's photo by exploiting state-of-the-art computer vision methods. Furthermore, a more general measure of cross-modal similarity of the textual content to the scene depicted in the image has been introduced. In contrast to previous work, the reference data for the visual representation of the extracted entities are not derived from a similar data source with additionally available metadata, but is fully-automatically obtained via large-scale image databases. Experiments were conducted on two datasets that contain real-world news articles across different topics, domains, and languages. The experimental results have clearly demonstrated the feasibility of the proposed approach. Datasets and source code are publicly available to further foster research towards this novel task of cross-modal consistency verification in news.

In the future, we aim to refine the queries based on the extracted text entities for the visual verification approach by further exploiting knowledge graph information. Another interesting direction of research is to investigate the impact of other entity types~(e.g., time), entity relations, as well as relations between the overall textual and visual sentiment to increase the performance of the system.
\section*{Acknowledgement}
This project has received funding from the European Union’s Horizon 2020 research and innovation programme under the Marie Skłodowska-Curie grant agreement no 812997, and the the German Research Foundation (DFG: Deutsche Forschungsgemeinschaft, project number: 388420599).
We also want to gratefully thank Avishek Anand~(L3S Research Center, Leibniz Universität Hannover) for his valuable comments to improve the paper.
% EOF

%% file: fig_intro.tex
\begin{figure}[t]
	\centering
    \includegraphics[width=1\linewidth]{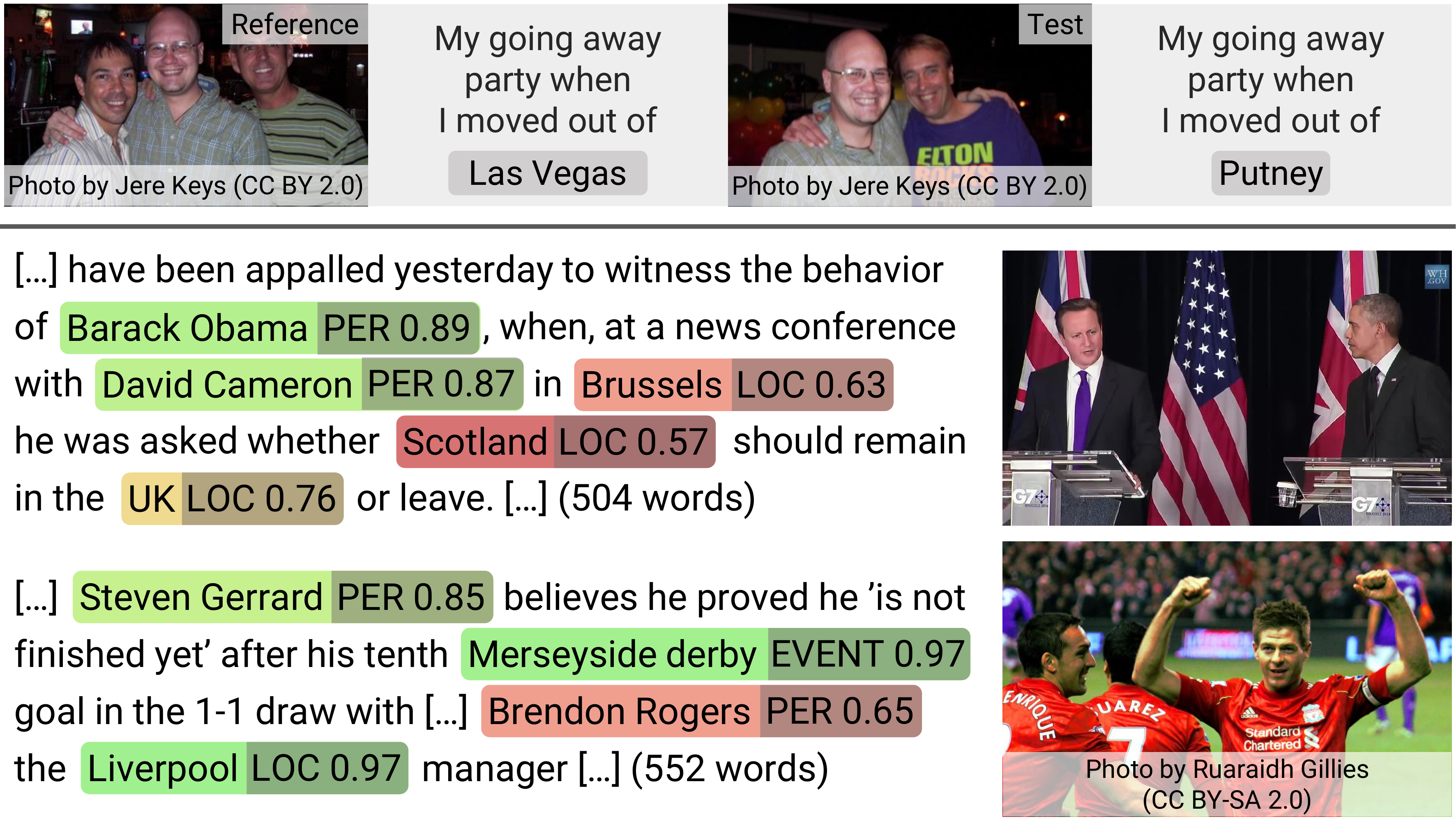}
	\caption{Top: Reference and test image of the \emph{MEIR} dataset~\cite{sabir2018deep} and corresponding texts with untampered and tampered entities. Bottom: Two real-world news from \emph{BreakingNews}~\cite{ramisa2018breakingnews} and outputs of our system~(LOC\-ation, PERson, EVENT). The examples show that real-world news have much longer text and refer to many entities. Images are replaced with similar ones due to licensing issues. Original images and full text are linked on the GitHub page\textsuperscript{\ref{foot:github}}.}% The original images can be viewed here: (a) \href{https://google.de}{News}, Image (b) News, Image}
	\label{fig:examples}
\end{figure}

%% file: tab_statistics_datasets.tex
\renewcommand{\b}{\textbf}
\begin{table*}[t]
    \setlength\tabcolsep{8pt}
    \caption{Number of test documents~$|D|$, unique entities~$T^\ast$ in all articles, and mean amount of unique entities~$\overline{T}$ in articles containing a given entity type~(for \emph{context} this is the mean amount of nouns as explained in Section~\ref{sec:txt_embedding}) for \emph{TamperedNews}~(left) and \emph{News400}~(right). Valid image-text relations for \emph{News400} were first manually verified according to Section~\ref{sec:news400}.} %and that events cannot be detected by \emph{spaCy} for $<$double-blind language$>$.}
    \label{tab:statistics_datasets}
    \centering
    \parbox{.48\linewidth}{
    \centering
    \begin{tabular}{l | c c c}
        \multicolumn{4}{c}{\b{TamperedNews dataset}} \\
        \b{Documents}                                   & $|D|$	    & $T^\ast$  & $\overline{T}$\\
        \hline
        All~(context)                                   & 72,561    & ---	    & 121.40    \\
        \hline
        With persons                                    & 34,051    & 4,784	    & 4.03      \\
        With locations                                  & 67,148    & 3,455	    & 4.90      \\
        With events                                     & 16,786    & 897	    & 1.33      \\
        
    \end{tabular}
    }
    \hfill
    \parbox{.5\linewidth}{
    \centering
%    \begin{tabular}{l | c c c}
%        \multicolumn{4}{c}{\b{News400 dataset}} \\
%        \b{Documents}                                   & $|D|$	    & $T^\ast$  & $\overline{T}$\\
%        \hline
%        % All                                             & 400       & ---	    & ---       \\
%		% \hline
%		With valid context        			            & 91        & ---       & 137.35   \\
%        \hline
%        With valid persons                              & 116       & 424       & 5.41     \\
%        With valid locations                  	        & 69        & 451	    & 9.22     \\
%        With valid events                  	            & 31        & 39	    & 1.84     \\
%    \end{tabular}

\begin{tabular}{l | c c c}
    \multicolumn{4}{c}{\b{News400 dataset}} \\
    \b{Documents}                                   & $|D|$	    & $T^\ast$  & $\overline{T}$\\
    \hline
    % All                                             & 400       & ---	    & ---       \\
	% \hline
	All (verified context)  	                    & 400 (91)        & ---       & 137.35   \\
    \hline
    With persons (verified)                         & 322 (116)       & 424       & 5.41     \\
    With locations (verified)              	        & 389 (69)        & 451	    & 9.22     \\
    With events (verified)            	            & 170 (31)        & 39	    & 1.84     \\
\end{tabular}

}
\end{table*}

% Breaking News: len(reference_images) in wd_{entity_type}-relevant (std)

% overall
%ALL: 38.25 (2.85)
%PLACES: 38.34 (2.73)
%PERSONS: 38.00 (2.88)
%EVENTS: 39.28 (2.91)

% per search engine
% PLACES - google: 18.68 (1.56)
% PLACES - bing: 18.78 (2.11)
% PLACES - wikidata: 0.88 (0.72)
% PERSONS - google: 18.26 (2.01)
% PERSONS - bing: 18.87 (2.05)
% PERSONS - wikidata: 0.88 (0.39)
% EVENTS - google: 19.33 (1.40)
% EVENTS - bing: 19.37 (2.31)
% EVENTS - wikidata: 0.58 (0.67)

% News400
%PLACES: 38.92 (3.13)
%PERSONS: 38.85 (2.83)
%EVENTS: 38.79 (3.11)
%ALL: 38.88 (3.00)
%PLACES - google: 18.90 (1.60)
%PLACES - bing: 19.11 (2.51)
%PLACES - wikidata: 0.90 (0.70)
%PERSONS - google: 18.88 (1.47)
%PERSONS - bing: 19.11 (2.33)
%PERSONS - wikidata: 0.85 (0.44)
%EVENTS - google: 19.22 (1.57)
%EVENTS - bing: 18.96 (2.55)
%EVENTS - wikidata: 0.62 (0.60)

%% file: fig_cms_small.tex
\begin{figure*}[t]
    \centering
    \begin{minipage}[b]{0.32\textwidth}
        \centering
        \includegraphics[width=\textwidth,trim={0.35cm 0.35cm 0.35cm 0.35cm},clip]{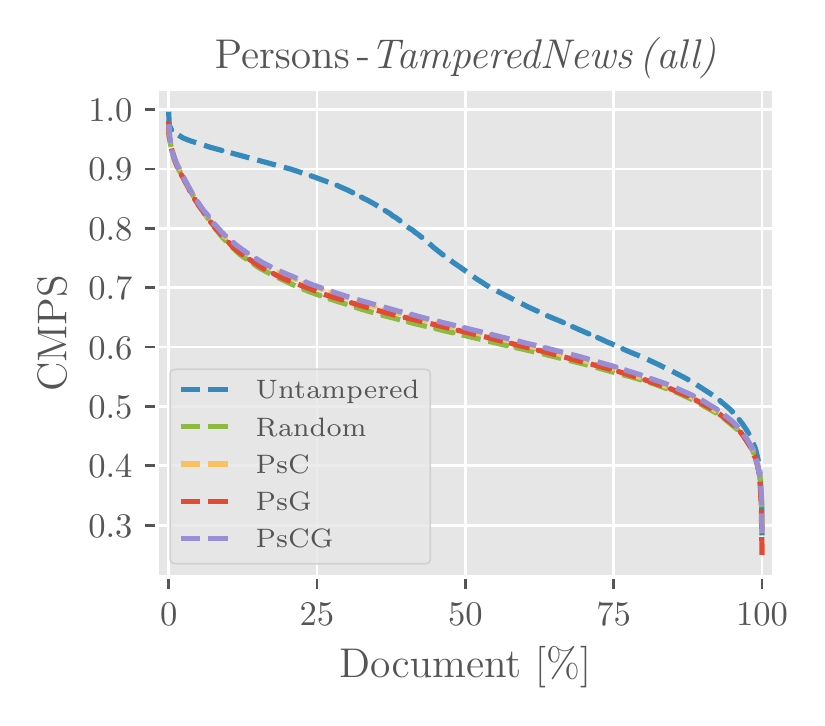}
    \end{minipage}
    \hfill
    \begin{minipage}[b]{0.32\textwidth}
        \centering
        \includegraphics[width=\textwidth,trim={0.35cm 0.35cm 0.35cm 0.35cm},clip]{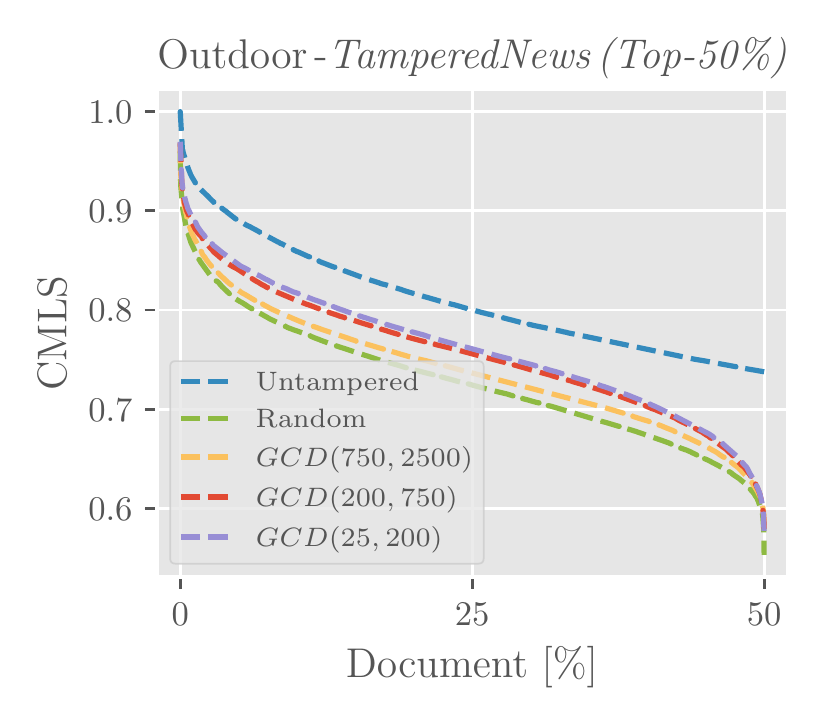}
    \end{minipage}
    \hfill
    \begin{minipage}[b]{0.32\textwidth}
        \centering
        \includegraphics[width=\textwidth,trim={0.35cm 0.35cm 0.35cm 0.35cm},clip]{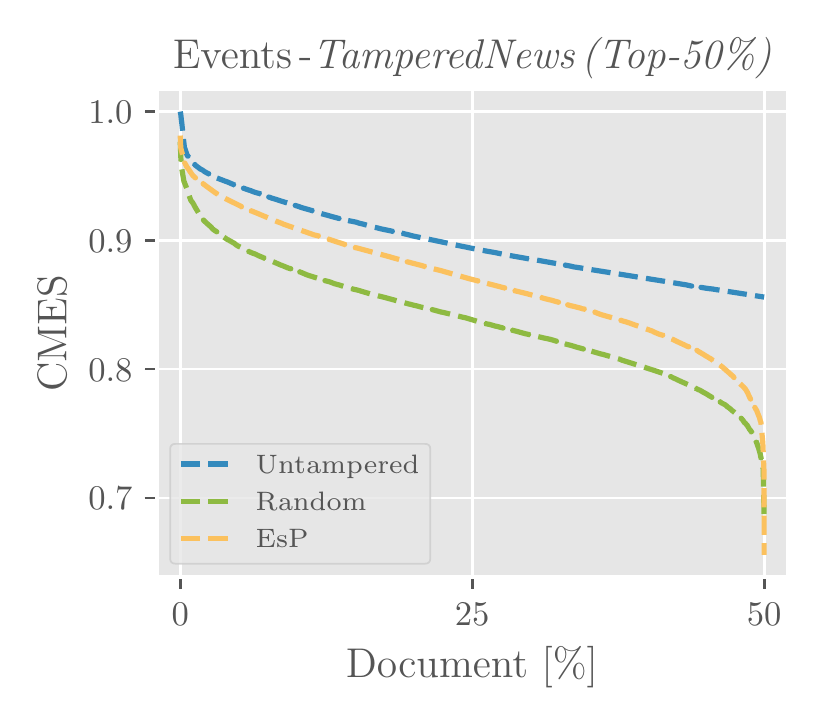}
    \end{minipage}
    \caption{Cross-modal similarity values of all~(or a subset of) documents sorted in descending order for person, location~(outdoor), and event entities for different tampering techniques~(Section~\ref{sec:tampering}) evaluated on the \emph{TamperedNews} dataset.} %The decision threshold~(red) is used for classification. Best viewed in color. The green area represents the Top-25\% documents with respect to the similarity.}
    \label{fig:cms}
\end{figure*}

%% file: tab_mode_results.tex
\begin{table}[t]
    \renewcommand{\b}{\textbf}
    \fontsize{8.5}{8.5}\selectfont
    \setlength\tabcolsep{3pt}
    \caption{AUC for different operators to calculate the cross-modal similarity for each entity of a given type~(Section~\ref{sec:align_features}) within a document. Results are reported for the \emph{TamperedNews~(Top-50\%)} validation dataset with the hardest tampering strategy~(notations according to Section~\ref{sec:tampering}).}
    \label{tab:mode_results}
    \centering
    \begin{tabular}{l | c | c | c c c c}
        \b{Test set}                        & |D|   & clustering & $Q_{75}$ & $Q_{90}$ & $Q_{95}$ & $\max$\\ 
        \hline
        Persons: PsCG                       & 1,703 & 0.94 & 0.92 & 0.95 & 0.95 & 0.89 \\
        Loc.-Outdoor: GCD$_{25}^{200}$      & 1,420 & ---  & 0.67 & 0.69 & 0.68 & 0.71 \\
        Loc.-Indoor: GCD$_{25}^{200}$       & 1,973 & ---  & 0.62 & 0.65 & 0.66 & 0.68 \\
        Events: EsP                         & 839   & ---  & 0.65 & 0.68 & 0.68 & 0.69 \\
    \end{tabular}
\end{table}

%% file: fig_results.tex
\begin{figure*}[t]
	\centering
    \includegraphics[width=0.9705\linewidth]{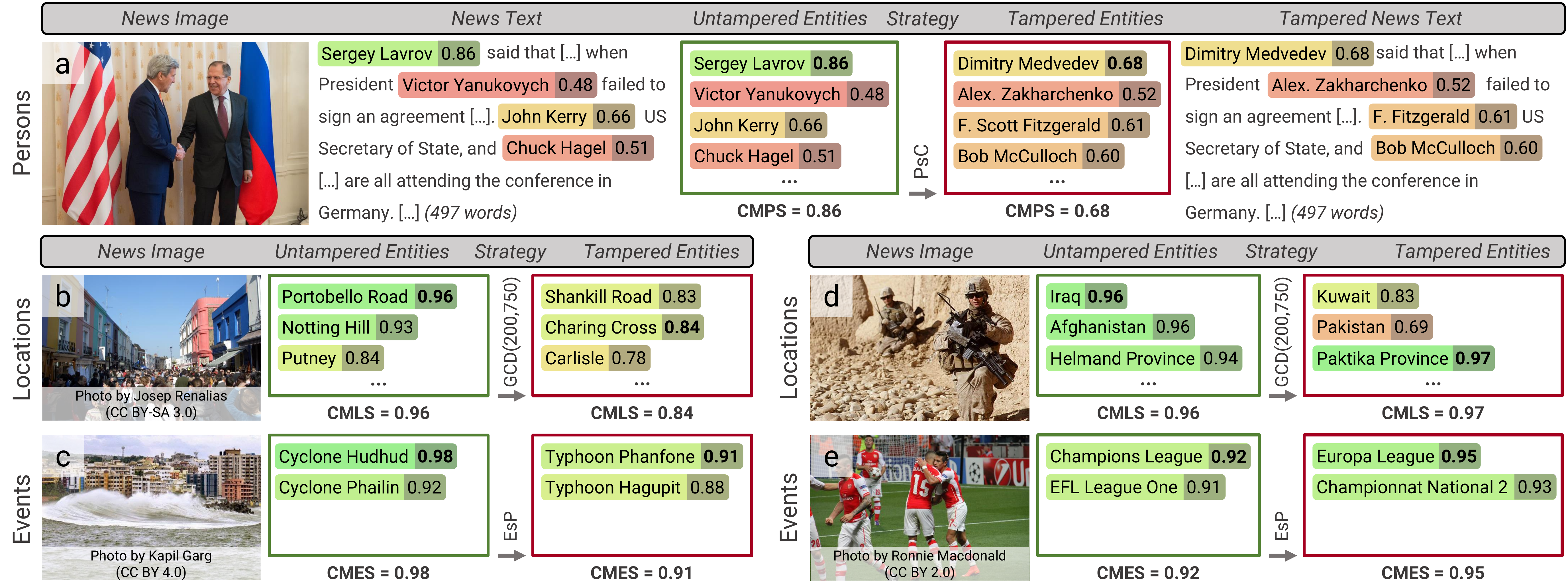}
	\caption{Positive (a-c, CMS of the untampered entity set is higher) and negative (d-e, CMS of the tampered entity set is higher) verification results of some \emph{TamperedNews} documents. Within each example the similarities~(from red to green with intervals: persons [0.45, 1], locations [0.6, 1], events [0.7, 1]) of the news image to a set of untampered entities~(green border) and tampered entities~(red border) using \emph{one} specific tampering strategy are shown. Images are replaced with similar ones due to licensing issues. Original images and full text are linked on the GitHub page\textsuperscript{\ref{foot:github}}.}
	\label{fig:case_study}
\end{figure*}

%% file: tab_results_breakingnews.tex
\renewcommand{\b}{\textbf}
\begin{table}[t]
% \caption{AUC, FRP, AP-clean and AP-tampered at different recall levels of the \emph{TamperedNews~(Top-50\%)} dataset for different entity testsets~(notations according Section~\ref{sec:tampering}.)}
\caption{Results for document verification~(DV) and collection retrieval for the \emph{TamperedNews~(Top-50\%)} dataset for different entity testsets~(notations according Section~\ref{sec:tampering}.)}
\label{tab:results_breakingnews}
\fontsize{8}{8}\selectfont

\setlength\tabcolsep{1pt}
\centering
    \begin{tabular}{l | c | c | c c c | c c c}

        \multirow{3}{*}{\b{Test set ($|D|$)}} & \b{DV} & \multicolumn{7}{c}{\b{Collection Retrieval}} \\
        & \multirow{2}{*}{\b{VA}} & \multirow{2}{*}{\b{AUC}} & \multicolumn{3}{c|}{\b{AP-clean}} & \multicolumn{3}{c}{\b{AP-tampered}}  \\
        & & & @25\%     & @50\%    & @100\%     & @25\%   & @50\%    & @100\%  \\
        
        \hline
        \b{Persons} (15,323)                    &&&&&&&\\
        $\;$ Random                            & 0.94 & 0.94 & 96.27 & 95.49 & 92.27 & 100.0 & 100.0 & 95.82 \\
        $\;$ PsC                               & 0.93 & 0.94 & 95.30 & 94.45 & 91.19 & 100.0 & 100.0 & 95.23 \\
        $\;$ PsG                               & 0.93 & 0.94 & 95.59 & 94.89 & 91.76 & 100.0 & 100.0 & 95.61 \\
        $\;$ PsCG                              & 0.93 & 0.94 & 94.94 & 94.54 & 91.17 & 100.0 & 100.0 & 95.13 \\
        
        \hline
        
        \b{Locations} (30,217)        &&&&&&&\\
        \textbullet$\,$\b{Outdoor} (12,780)           &&&&&&&\\
        $\;$ Random                            & 0.87 & 0.85 & 92.28 & 87.79 & 81.29 & 100.0 & 100.0 & 88.42 \\
        $\;$ GCD(750, 2500)                    & 0.86 & 0.81 & 88.22 & 83.81 & 77.24 & 100.0 & 100.0 & 85.30 \\
        $\;$ GCD(200, 750)                     & 0.79 & 0.74 & 83.42 & 77.09 & 70.44 & 100.0 & 96.57 & 78.98 \\
        $\;$ GCD(25, 200)                      & 0.76 & 0.71 & 80.36 & 73.88 & 67.89 & 100.0 & 94.62 & 76.91 \\
        
        %$\;$\textbullet$\,$\b{Indoor} (19,372)       &&&&&&&\\
        \textbullet$\,$\b{Indoor} (17,437)                     &&&&&&&\\
        $\;$ Random                            & 0.74 & 0.72 & 69.68 & 67.26 & 65.23 & 100.0 & 97.77 & 78.56 \\
        $\;$ GCD(750, 2500)                    & 0.71 & 0.68 & 63.20 & 62.39 & 61.79 & 100.0 & 95.32 & 76.09 \\
        $\;$ GCD(200, 750)                     & 0.72 & 0.69 & 67.69 & 65.39 & 63.50 & 100.0 & 95.75 & 76.68 \\
        $\;$ GCD(25, 200)                      & 0.69 & 0.67 & 57.46 & 58.56 & 59.39 & 100.0 & 93.91 & 74.75 \\
        
        \hline
        
        \b{Events} (7,554)                      &&&&&&&\\
        $\;$ Random                            & 0.88 & 0.87 & 91.11 & 88.26 & 83.09 & 100.0 & 100.0 & 90.50 \\
        $\;$ EsP                               & 0.75 & 0.70 & 69.61 & 66.42 & 64.09 & 100.0 & 96.17 & 76.95 \\
        
        \hline
        
        \b{Context} (36,217)                    &&&&&&&\\
        $\;$ Random                            & 0.81 & 0.80 & 88.95 & 83.03 & 76.32 & 100.0 & 100.0 & 84.79 \\
        $\;$ Similar~(top-$25\%$)              & 0.78 & 0.77 & 83.52 & 78.12 & 72.43 & 100.0 & 99.70 & 82.25 \\
        $\;$ Similar~(top-$10\%$)              & 0.76 & 0.74 & 77.76 & 73.21 & 68.78 & 100.0 & 98.33 & 79.84 \\
        $\;$ Similar~(top-$5\%$)               & 0.74 & 0.71 & 74.31 & 69.89 & 66.22 & 100.0 & 96.84 & 77.92 \\
    \end{tabular}
\end{table}

%% file: tab_meta_results.tex
\renewcommand{\b}{\textbf}
\begin{table*}[t]
    \setlength\tabcolsep{4pt}
    \caption{AUC for a selection of location~(left) and event~(right) types of the \emph{TamperedNews~(Top-50\%)} dataset. $|D|$~is the number of documents containing at least one entity of this type and $|D_s|$~is the number of times this type has achieved the highest cross-modal similarity within the untampered set. Results are reported considering the documents~$D_s$ for each entity type.}
    \label{tab:meta_results}
    \fontsize{8}{8}\selectfont
    \centering
    \parbox{.56\linewidth}{
    \centering
    \begin{tabular}{l | c c | c c c c}
        \multicolumn{7}{c}{\b{Selection of 12 / 1,067 location entity types}} \\
        % \multicolumn{7}{c}{}\\
        \multirow{2}{*}{\b{Type} (num. of entities)} & \multirow{2}{*}{$|D|$} & \multirow{2}{*}{$|D_s|$} & \multicolumn{4}{c}{\b{AUC}} \\
        & & & Random & GCD$^{2500}_{750}$ & GCD$^{750}_{200}$ & GCD$^{200}_{25}$ \\
        \hline
continent (7) & 1,692 & 112 & 0.83 & 0.68 & 0.71 & 0.70 \\
country (184) & 8,689 & 2,579 & 0.85 & 0.75 & 0.71 & 0.69 \\
% island (96) & 4,139 & 583 & 0.84 & 0.75 & 0.73 & 0.69 \\
state (108) & 1,802 & 337 & 0.85 & 0.71 & 0.68 & 0.68 \\
\hline
city (726) & 8,912 & 2,821 & 0.84 & 0.83 & 0.76 & 0.73 \\
town (604) & 4,508 & 1,501 & 0.79 & 0.87 & 0.70 & 0.67 \\
district (65) & 1,684 & 299 & 0.81 & 0.86 & 0.79 & 0.73 \\
street (25) & 313 & 58 & 0.80 & 0.72 & 0.74 & 0.72 \\
tourist attraction (64) & 752 & 139 & 0.94 & 0.91 & 0.88 & 0.88 \\
% building (42) & 551 & 71 & 0.86 & 0.82 & 0.82 & 0.84 \\
\hline
mountain range (13) & 181 & 43 & 0.91 & 0.81 & 0.65 & 0.62 \\
mountain (9) & 77 & 29 & 0.96 & 0.89 & 0.82 & 0.80 \\
\hline
ocean (4) & 330 & 49 & 0.91 & 0.51 & 0.46 & 0.52 \\
% sea (14) & 386 & 96 & 0.88 & 0.86 & 0.81 & 0.80 \\
river (42) & 418 & 113 & 0.84 & 0.80 & 0.72 & 0.69 \\

    \end{tabular}
    }
    \hfill
    \parbox{.39\linewidth}{
    \centering
    \begin{tabular}{l | c c | c c}
        \multicolumn{5}{c}{\b{Selection of 12 / 493 event entity types}} \\
        % \multicolumn{5}{c}{}\\
        \multirow{2}{*}{\b{Type} (num. of entities)} & \multirow{2}{*}{$|D|$} & \multirow{2}{*}{$|D_s|$} & \multicolumn{2}{c}{\b{AUC}} \\
         &  &  & Random & EsP \\
        \hline
competition (32) & 1,371 & 1,150 & 0.92 & 0.68 \\
sport competition (16) & 144 & 90 & 0.86 & 0.61 \\
\hline
festival (72) & 443 & 351 & 0.90 & 0.81 \\
award (6) & 346 & 281 & 0.81 & 0.70 \\
% ceremony (4) & 60 & 35 & 0.83 & 0.70 \\
holiday (30) & 261 & 110 & 0.87 & 0.91 \\
convention (8) & 61 & 56 & 0.81 & 0.70 \\
\hline
war (44) & 758 & 491 & 0.82 & 0.75 \\
shooting (6) & 72 & 61 & 0.83 & 0.75 \\
% terrorist attack (4) & 30 & 20 & 0.89 & 0.76 \\
% accident (6) & 105 & 102 & 0.93 & 0.87 \\
disaster (6) & 62 & 53 & 0.97 & 0.94 \\
\hline
scandal (10) & 103 & 92 & 0.90 & 0.64 \\
legal case (10) & 43 & 42 & 0.89 & 0.77 \\
protest (9) & 60 & 50 & 0.94 & 0.72 \\

    \end{tabular}
}
\end{table*}

%% file: tab_results_news400.tex
\renewcommand{\b}{\textbf}
\begin{table}[t]
\caption{Results for document verification~(DV) and collection retrieval for the the \emph{News400} dataset. Results are reported for all available and verified documents~$|D|$.} %for different entity types and tampering methods.}
\label{tab:results_news400}
\fontsize{8}{8}\selectfont
%\fontsize{8}{8.8}\selectfont
\setlength\tabcolsep{1pt}
\centering
    \begin{tabular}{l | c | c | c c c | c c c}
       
        \multirow{3}{*}{\b{Test set ($|D|$)}} & \b{DV} & \multicolumn{7}{c}{\b{Collection Retrieval}} \\
        & \multirow{2}{*}{\b{VA}} & \multirow{2}{*}{\b{AUC}} & \multicolumn{3}{c|}{\b{AP-clean}} & \multicolumn{3}{c}{\b{AP-tampered}}  \\
        & & & @25\%     & @50\%    & @100\%     & @25\%   & @50\%    & @100\%  \\
        
        \hline
        
        \b{Persons} (116)                               &&&&&&&\\
        $\;$ Random                                    & 0.94 & 0.92 & 100.0 & 100.0 & 93.79 & 87.68 & 87.57 & 86.77 \\
        $\;$ PsC                                       & 0.93 & 0.90 & 100.0 & 99.49 & 92.30 & 83.77 & 84.91 & 84.48 \\
        $\;$ PsG                                       & 0.91 & 0.91 & 98.95 & 98.24 & 92.29 & 82.80 & 84.86 & 84.93 \\
        $\;$ PsCG                                      & 0.93 & 0.91 & 100.0 & 99.82 & 93.62 & 86.63 & 86.66 & 86.15 \\
        
        \hline
        \b{Location} (69)                       &&&&&&&\\
        \textbullet$\,$\b{Outdoor} (54)                       &&&&&&&\\
        $\;$ Random                                    & 0.89 & 0.85 & 100.0 & 98.44 & 88.21 & 82.91 & 81.36 & 79.91 \\
        $\;$ GCD(750, 2500)                            & 0.81 & 0.80 & 92.61 & 88.51 & 81.03 & 68.77 & 70.84 & 72.64 \\
        $\;$ GCD(200, 750)                             & 0.80 & 0.75 & 87.55 & 81.95 & 74.77 & 65.07 & 68.38 & 68.34 \\
        $\;$ GCD(25, 200)                              & 0.81 & 0.73 & 87.55 & 81.72 & 73.33 & 66.17 & 70.16 & 68.08 \\
                
        \color[gray]{0.5}\textbullet$\,$\b{Indoor} (15)       &&&&&&&&\\
        \color[gray]{0.5}$\;$ Random                   & \color[gray]{0.5}0.80 & \color[gray]{0.5}0.76& \color[gray]{0.5}91.67& \color[gray]{0.5}81.98& \color[gray]{0.5}76.07& \color[gray]{0.5}88.75& \color[gray]{0.5}86.44& \color[gray]{0.5}77.94\\
        \color[gray]{0.5}$\;$ GCD(750, 2500)           & \color[gray]{0.5}0.67 & \color[gray]{0.5}0.64& \color[gray]{0.5}62.20& \color[gray]{0.5}59.98& \color[gray]{0.5}60.52& \color[gray]{0.5}80.42& \color[gray]{0.5}81.06& \color[gray]{0.5}69.09\\
        \color[gray]{0.5}$\;$ GCD(200, 750)            & \color[gray]{0.5}0.87 & \color[gray]{0.5}0.70& \color[gray]{0.5}73.33& \color[gray]{0.5}69.10& \color[gray]{0.5}67.56& \color[gray]{0.5}67.92& \color[gray]{0.5}73.47& \color[gray]{0.5}69.10\\
        \color[gray]{0.5}$\;$ GCD(25, 200)             & \color[gray]{0.5}0.73 & \color[gray]{0.5}0.66& \color[gray]{0.5}74.70& \color[gray]{0.5}68.95& \color[gray]{0.5}65.66& \color[gray]{0.5}67.92& \color[gray]{0.5}70.32& \color[gray]{0.5}65.29\\
        
        \hline
        \b{Events} (31)                                 &&&&&&&\\
        $\;$ Random                                    & 0.87 & 0.85 & 92.94 & 89.10 & 83.37 & 100.0 & 92.44 & 85.28 \\
        $\;$ EsP                                       & 0.65 & 0.64 & 52.76 & 56.43 & 58.38 & 85.10 & 80.71 & 68.98 \\
        
        \hline
        
        \b{Context} (91)                          &&&&&&&\\
        $\;$ Random                                    & 0.70 & 0.70 & 87.03 & 87.50 & 73.62 & 61.11 & 63.09 & 63.19 \\
        $\;$ Similar~(top-$25\%$)                      & 0.70 & 0.68 & 92.19 & 88.43 & 72.96 & 53.60 & 57.77 & 59.69 \\
        $\;$ Similar~(top-$10\%$)                      & 0.64 & 0.66 & 70.54 & 74.12 & 65.58 & 56.15 & 59.72 & 59.75 \\
        $\;$ Similar~(top-$5\%$)                       & 0.66 & 0.63 & 74.48 & 73.09 & 64.18 & 50.77 & 55.99 & 56.98 \\
    \end{tabular}
\end{table}